\documentclass[11pt]{article}

\usepackage{amsmath}
\usepackage[final]{acl}
\usepackage[utf8]{inputenc}
\usepackage{CJKutf8}
\usepackage{authblk}
\usepackage{xspace}
\usepackage{caption}
\usepackage{mdframed}
\usepackage{times}
\usepackage{latexsym}
\usepackage[utf8]{inputenc}
\usepackage[T1]{fontenc}
\usepackage[utf8]{inputenc}

\usepackage{microtype}
\usepackage{multirow}

\usepackage{inconsolata}

\usepackage{graphicx}
\title{BaZi-Based Character Simulation Benchmark: Evaluating AI on Temporal and Persona Reasoning}

\author{
Siyuan Zheng$^{* \clubsuit \triangle}$, 
Pai Liu$^{*\dagger \clubsuit \spadesuit}$,
Xi Chen$^{* \clubsuit \spadesuit}$,
Jizheng Dong$^{\heartsuit}$,
Sihan Jia$^{\diamondsuit}$
}

\affil{
$^{\clubsuit}$MirrorAI Co., Ltd. \quad
$^{\spadesuit}$University of Rochester \quad
$^{\heartsuit}$New York University \\
$^{\diamondsuit}$Georgia State University \quad
$^{\triangle}$Anhui Zhu Zi College \\
\texttt{\{pi, siyuanz\}@mymirrorai.com}
}

\begin{document}
\begin{CJK}{UTF8}{gbsn}
\maketitle

\begingroup
\renewcommand\thefootnote{}%
\footnotetext{%
\parbox{\linewidth}{
\hspace{-5mm}$^{*}$Equal contribution 

\hspace{-5mm}$^{\dagger}$The corresponding author
}%
}
\endgroup

\begin{abstract}
Human-like virtual characters are crucial for games, storytelling, and virtual reality, yet current methods rely heavily on annotated data or handcrafted persona prompts, making it difficult to scale up and generate realistic, contextually coherent personas. We create the first QA dataset for BaZi-based persona reasoning, where real human experiences categorized into wealth, health, kinship, career, and relationships are represented as life-event questions and answers. Furthermore, we propose the first BaZi–LLM system that integrates symbolic reasoning with large language models to generate temporally dynamic and fine-grained virtual personas. Compared with mainstream LLMs such as DeepSeek-v3 and GPT-5-mini, our method achieves a \textbf{30.3\%–62.6\% accuracy improvement}. In addition, when incorrect BaZi information is used, our model's accuracy drops by 20\%–45\%, showing the potential of culturally grounded symbolic–LLM integration for realistic character simulation.
\end{abstract}

\section{Introduction}
\vspace{-2mm}
The development of realistic virtual characters is central to immersive applications in gaming, storytelling, and interactive media. Traditional approaches such as dialogue trees, finite-state machines, and behavior trees—are costly to author, brittle beyond narrow scenarios, and tend to yield template-like personas with weak long-horizon consistency \cite{millington2019aiforgames, colledanchise2018behaviortrees}. Large Language Models (LLMs) such as DeepSeek, Qwen, and ChatGPT have demonstrated strong prompt-following and dialogue generation \cite{brown2020language, openai2023gpt4}, enabling LLM-based NPCs, generative agents \cite{park2023generativeagents}, and multi-agent simulations \cite{wang2023voyager}. Yet these pipelines still face limitations: detailed persona prompts cannot capture human complexity within length constraints \cite{liu2023pretrainpromptpredict}, and character-specific finetuning is difficult to scale across diverse personas \cite{hu2022lora, dettmers2023qlora}.

\begin{figure}
    \centering
    \includegraphics[width=1.0\linewidth]{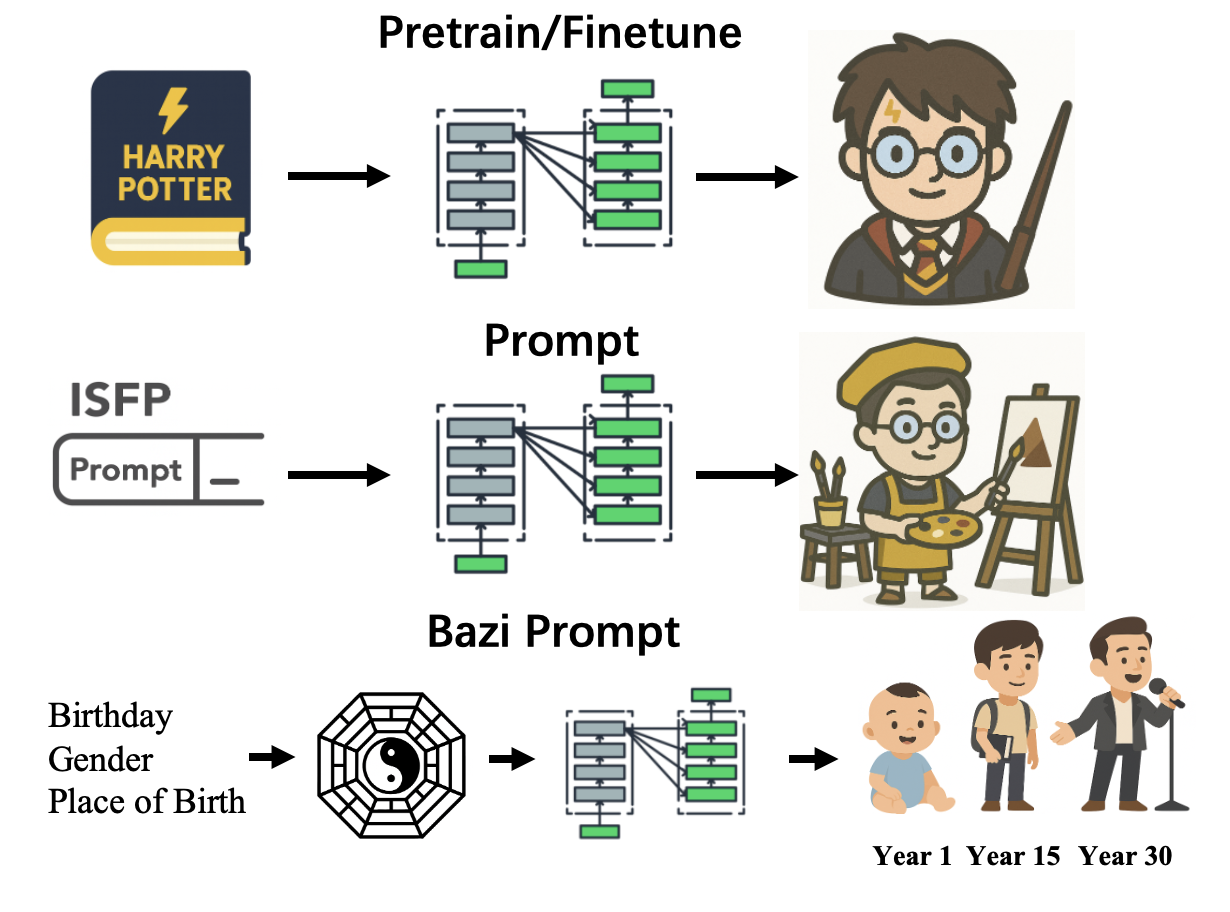}
    \vspace{-2mm}
    \caption{Mainstream approaches to character simulation rely on either pretraining/finetuning on existing literary works or prompt-based conditioning. In contrast, we propose a novel Bazi-prompt framework, which encodes birth information (birthday, gender, and place of birth) into symbolic features. This framework enables finer-grained character simulation in terms of personality, temporal dynamics, and more diverse interactions with different environments.}
    \label{fig:intro}
\end{figure}  

Motivated by these gaps, we adopt BaZi (the Four Pillars of Destiny) as a culturally grounded, temporally structured representation for persona construction as shown in Figure \ref{fig:intro}. 
BaZi encodes birth time into eight structured symbols (Heavenly Stems/Earthly Branches), provides mappings between personality facets and person–environment interactions, and offers temporal dynamics via Flowing Years, Months and Days\footnote{https://en.wikipedia.org/wiki/Four\_Pillars\_of\_Destiny}. Similar to how astrology or MBTI serve as cultural vocabularies \cite{campion2012popular, furnham1996mbti}, BaZi functions here as a narrative representation for identity and life-course description \cite{homola2021eightsigns} in an interpretable, probabilistic manner. 
While the full BaZi system also considers postnatal life events, acknowledging that later experiences can partially influence future trajectories rather than being solely determined by birth information (analogous to probabilistic variations within a 3-sigma range), this work focuses on the fundamental factors of birth time, place of birth, and gender to simplify modeling and capture the core generative components, while recognizing that this abstraction inevitably reduces theoretical fidelity.
We reinterpret BaZi as a conditional feature-generation model that discretizes chronological time into symbolic attributes tied to personal traits and temporal dynamics, enabling fine-grained, dynamic persona generation without metaphysical claims.

\subsection*{Empirical Motivation for BaZi-Based Temporal Modeling}

A substantial empirical literature demonstrates that conventional temporal markers related to birth timing correlate with important life outcomes. In education, season-of-birth and school-entry relative age effects explain measurable differences in achievement (e.g., \citet{crawford2014drivers}; \citet{field2015season}). In health, large phenome-wide association studies find systematic birth-month associations with risks for respiratory and cardiovascular conditions with independent replications (e.g., \citet{boland2015phenome}; \citet{li2016replication}). Biological mechanisms have also been proposed linking prenatal timing to later-life physiology (e.g., \citet{hemati2021systematic}; \citet{disanto2012month}). 
This contrasts with the lack of predictive validity for astrological indicators (e.g., \citet{carlson1985doubleblind}; \citet{wyman2008science}; \citet{deanMetaAstrology}) motivates our design: we establish state-of-the-art LLMs as strong Temporal Baseline using empirically supported variables (birth date and time), and then assess whether BaZi-derived symbolic features provide an incremental signal for persona reasoning beyond these conventional temporal effects.

We compare our proposed BaZi-augmented model with state-of-the-art LLMs, including Gemini-2.5-Flash, DeepSeek-v3, and GPT-5-Mini, on the Celebrity~50 dataset for life-event prediction. Our model achieves accuracy improvements of 30.3\% over DeepSeek-v3 and 62.6\% over GPT-5-Mini. To evaluate the impact of incorporating the BaZi system, we further compare model performance with and without shuffled personal profiles. When the mappings between individuals and their questions are randomized, our model’s performance drops by up to 45.7\%. These results demonstrate the effectiveness of the BaZi system in enhancing persona generation.
 
\subsection*{Dataset Design: QA over Life Events}
Evaluating character simulation via full life-course narratives is inherently difficult to verify. We therefore create a QA-based dataset, \textit{Celebrity 50}, focused on critical life events and containing information about 50 real individuals from diverse global backgrounds. Each persona is associated with 4–5 question–answer pairs spanning five key life stages (wealth, health, kinship, career, and relationships). This formulation reduces evaluation complexity while enabling reasoning over significant, discrete life nodes. It aligns with existing benchmarks in reasoning and commonsense evaluation \cite{talmor2019commonsenseqa}, enables quantitative evaluation through structured question–answer pairs, addressing a long-standing limitation of BaZi reasoning, which previously lacked measurable accuracy.

\subsection*{Our Contributions}
In summary, our main contributions are as follows:
(1) We reinterpret BaZi as a culturally grounded representational system for persona simulation, enabling fine-grained and temporally dynamic character modeling;
(2) We create the first QA dataset for BaZi-based persona reasoning, allowing systematic and quantitative evaluation of symbolic reasoning in life-event contexts;
(3) We develop the first BaZi-augmented system that integrates symbolic reasoning with LLMs for culturally informed character simulation;
(4) Our BaZi-enhanced models achieve consistent accuracy gains over baseline LLMs across all tested backbones on the Celebrity~50 benchmark.  

\section{Datasets}
Our multilingual dataset \textbf{Celebrity 50} is designed to evaluate Large Language Models' (LLMs) ability to predict key life events. It features a primary collection from multiple countries for diversity and comprehensiveness. We collected and validated biographical records for 50 modern figures through \textit{astro.com}, restricting the selection to individuals born around 1940 to ensure sufficiently rich data.

\noindent\textbf{Data Filtering and Selection Criteria}~~A rigorous filtering process established four criteria for subject selection: they must be adults with sufficiently rich life experiences, excluding idols for privacy, and all must be born in the Northern Hemisphere. Based on this, we focused on 50 well-known individuals from various Northern Hemisphere countries born around 1940 to ensure adequate biographical data and diversity.

\begin{mdframed}[linewidth=1pt]
\setlength{\parindent}{0pt}










\textbf{Example}

\textbf{Birth time}: 1966/10/18, 11:15 PM

\textbf{Gender}: Female

\textbf{Place of birth}: Hong Kong

\textbf{Questions}: What kind of job is this person likely to have?

A. Lawyer.

B. Salesperson.

C. Real estate business

D. Library clerk

\end{mdframed}
\captionof{figure}{Sample information input to LLM} \label{ex:career2005}

\begin{figure}[h]
  \centering
  \includegraphics[width=0.5\textwidth]{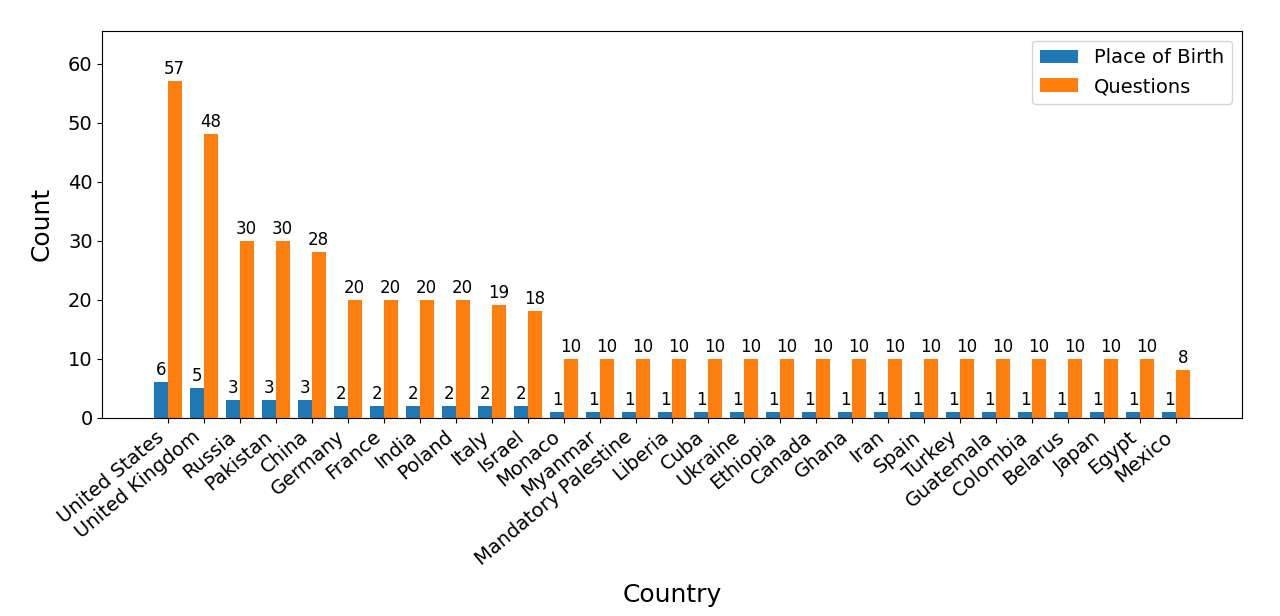}
  \caption{Question and Birthplace Counts Across Countries}
  \label{fig:country_analysis}
\end{figure}

\vspace{1em}
\noindent\textbf{Data Statistics}~~Comprehensive statistical analysis shows the dataset encompasses $\mathbf{50}$ individuals from $\mathbf{29}$ countries, totaling $\mathbf{488}$ question-answer pairs (avg. $\approx 9.76$ questions per person), as illustrated in Figure \ref{fig:country_analysis}. The gender distribution includes $\mathbf{37}$ males and $\mathbf{13}$ females, ensuring diverse demographic representation.

\vspace{1em}

\noindent\textbf{Construction Process}~~The annotation process starts with acquiring precise birth time data. The Qwen API is then prompted to retrieve biographical narratives across five dimensions—wealth, health, kinship, career, and relationships—leveraging its web search and internal knowledge base. The same LLM generates multiple-choice questions from this compiled information. A final script extracts and synthesizes these questions with the birth data into the target JSON format. The authors conduct the entire process, including data cleaning, filtering, and structuring.

\vspace{1em}

\noindent\textbf{Cleaning and Quality Assurance}~~The initial LLM-generated questions underwent a rigorous cleaning process involving both automated refinement and manual verification to ensure quality and appropriateness.

First, we established a rating system based on three criteria for elimination:
\begin{itemize}
    \item Questions containing real proper names (people, organizations, teams, etc.).
    \item Questions demanding overly specific numerical details (e.g., exact wealth amounts) that are not reasonably predictable by Bazi analysis.
    \item Questions that exceed the reasonable predictive capabilities of traditional Bazi analysis.
\end{itemize}
Unsatisfactory questions were grouped and iteratively refined by the LLM itself through prompt modifications. Discarded questions were replaced by new ones generated from updated prompts. Finally, all remaining questions underwent manual verification to ensure rigor and compliance with our guidelines.

\vspace{1em}

\noindent\textbf{Annotation Process}~~Comprehensive guidelines were developed for this task. The core requirement is that all generated questions must be factually accurate and strictly align with one of the five predefined life dimensions based on the sourced biographical material. The complete annotation guideline is provided in the Appendix for reference.

\noindent\textbf{Task Definition}~~The model's input is the individual's birth time, gender, and place of birth, along with multiple-choice questions (Figure \ref{ex:career2005}). The target output is the correct answer choice, which requires applying destiny analysis principles to the provided context\footnote{We have released our dataset, output cases and experiment details at https://github.com/MirrorAI-Lab/BaZi-Persona}.

\section{Method}
\label{sec:method}
\begin{figure}
    \centering
    \includegraphics[width=1\linewidth]{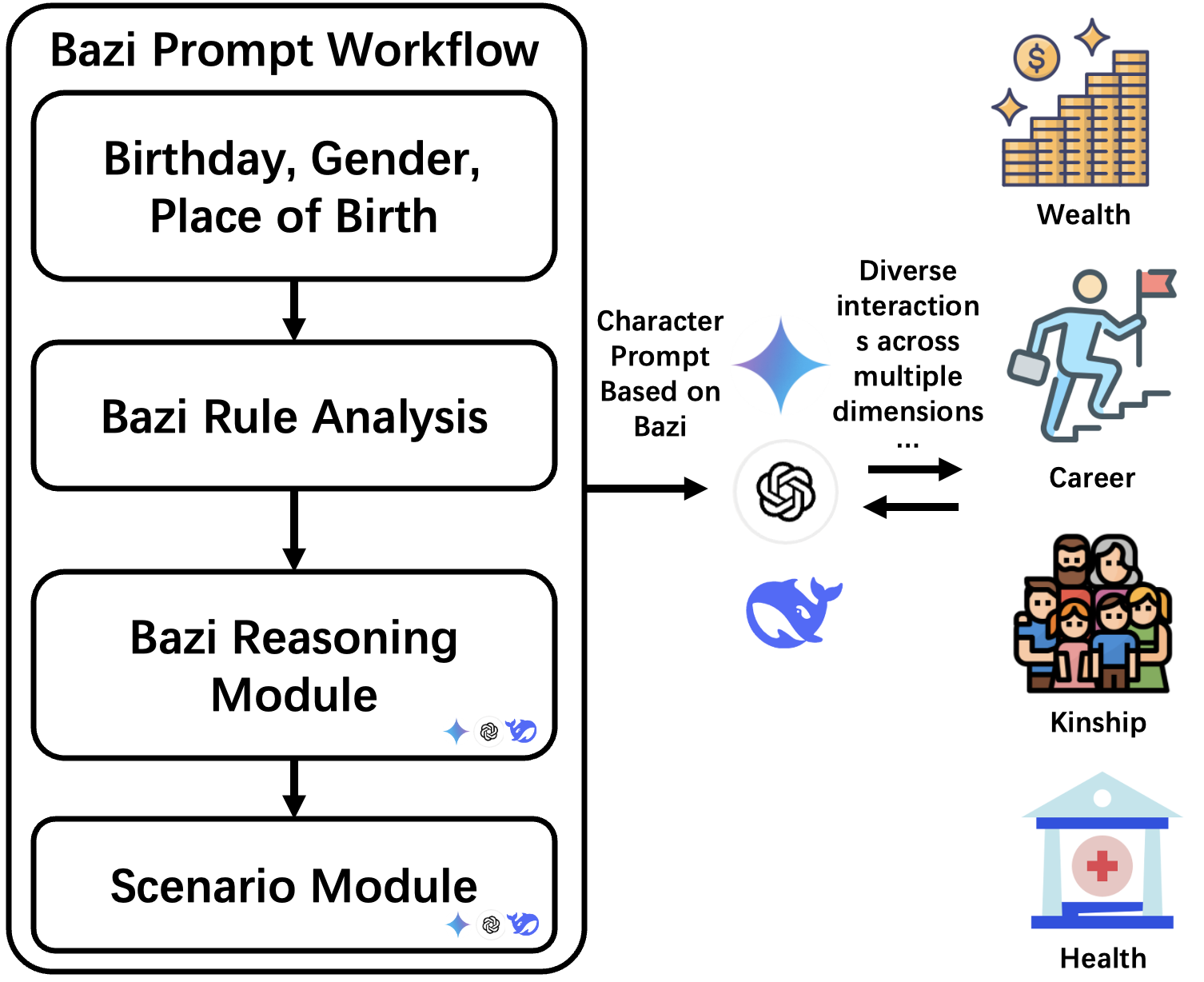}
    \caption{Our model is organized into four main components: (1) input layer for birth-related information (birthday, gender, place of birth), (2) BaZi rule analysis, (3) BaZi reasoning, and (4) scenario-specific interpretation. The BaZi-LLM prompt workflow outputs fine-grained features describing personality traits and dynamic states of daily interactions with external dimensions such as wealth, career, kinship, and health.}
    \label{fig:model}
\end{figure}
\subsection{Overview}
As shown in Figure \ref{fig:model}, we propose a BaZi-inspired character simulation framework that systematically transforms an individual's birth information into structured, interpretable prompts. Our method is rooted in the classical BaZi (Four Pillars of Destiny) system, which encodes birth \textbf{year, month, day, and time} into symbolic features. Instead of treating BaZi as a purely metaphysical practice, we reinterpret it as a \textbf{symbolic rule-mapping process} that yields fine-grained temporal attributes for persona construction.

The input to our model requires only three elements: \textbf{birth date and time, gender, and place of birth}. From these minimal inputs, our pipeline generates temporally dynamic persona prompts that capture both \textbf{stable personality traits} and \textbf{temporal states}.

\subsection{BaZi Rule Mapping}
In the first stage, we implement a rule-based mapping program grounded in BaZi theory. This module translates the birth information into a structured BaZi chart, which consists of eight symbolic elements (heavenly stems and earthly branches). Each symbolic element is further associated with attributes reflecting:
\begin{itemize}
    \item \textbf{Personality features:} derived from the balance of the Five Elements (Wood, Fire, Earth, Metal, Water) and their corresponding roles.
    \item \textbf{Daily dynamic states:} extracted temporal features linked to health, career, wealth, and kinship.
\end{itemize}
This symbolic mapping ensures that the generated features are interpretable, structured, and temporally grounded.

\subsection{Interpretation via Classical Logic}
While the BaZi chart provides raw symbolic features, effective persona construction requires interpretive reasoning. We therefore design a coarse-grained interpretation mechanism inspired by classical BaZi analysis, which incorporates:
\begin{itemize}
    \item \textbf{Ten Gods (十神):} symbolic roles representing relationships between the day master (self) and other stems/branches.
    \item \textbf{ShenSha (神煞):} auxiliary symbolic markers associated with specific life tendencies or external influences.
    \item \textbf{Pattern Structures (格局):} higher-level symbolic groupings that reflect broader personality orientations.
\end{itemize}
The interpretive process follows the logic of BaZi divination, but rather than delivering deterministic outcomes, it produces \textbf{conditional interpretive features} that serve as the foundation for downstream scenario reasoning.

\subsection{Scenario-Oriented Analysis}
To enhance granularity, we couple the BaZi-derived interpretive features with \textbf{scenario-specific modules}. These modules contextualize the symbolic features into five primary domains: \textbf{Health}, \textbf{Career}, \textbf{Wealth}, \textbf{Relationship}, and \textbf{Kinship}.

This stage enables adaptive persona modeling, where symbolic features interact with environment-specific events. For instance, a feature that indicates career ambition may manifest differently when the external scenario involves interpersonal conflict versus financial opportunity.

\subsection{Dynamic Persona Prompt Generation}
Finally, the interpreted features are consolidated into dynamic prompts that simulate individual behavior and responses across time. Unlike static personality labels, our prompts incorporate both \textbf{long-term stable traits} and \textbf{short-term temporal variations}, thereby yielding a \textbf{time-sequenced and environment-aware character profile}. These prompts serve as the basis for generating lifelike and context-sensitive character simulations.

Our approach introduces three key methodological innovations: (1) \textbf{Minimal Input, Rich Output:} The model requires only birth information (date/time, gender, place of birth) yet produces temporally dynamic and domain-specific persona prompts. (2) \textbf{Symbolic-Logical Integration:} By combining rule-based BaZi mapping with interpretive logic (Ten Gods, ShenSha, Pattern Structures), the model generates structured symbolic features with explicit interpretability. (3) \textbf{Scenario Adaptivity:} Persona representations are not fixed; they dynamically adapt to health, career, wealth, relationship and kinship contexts, resulting in vivid, time-evolving character simulation.


\section{Experiments}

\begin{table}[ht] 
\centering
\scriptsize 
\begin{tabular}{|l|l|c|}
\hline 
Setting & Model & Acc. (\%) \\
\hline 
\multirow{3}{*}{Vanilla LLM w/ Bazi (Baseline)} 
& Deepseek-v3 & 39.3 \\
\cline{2-3} 
& Gemini-2.5-flash & 42.2 \\
\cline{2-3} 
& GPT-5-mini & 34.0 \\
\hline 
\multirow{3}{*}{Baseline w/ Bazi Rule Knowledge} 
& Deepseek-v3 & 35.9 ($\downarrow$8.7\%) \\
\cline{2-3} 
& Gemini-2.5-flash & 42.4 ($\downarrow$4.1\%) \\
\cline{2-3} 
& GPT-5-mini & 36.9 ($\uparrow$8.5\%) \\
\hline 
\multirow{3}{*}{Our Model} 
& Deepseek-v3 & 51.2 ($\uparrow$30.3\%) \\
\cline{2-3} 
& Gemini-2.5-flash & 47.1 ($\uparrow$6.6\%) \\
\cline{2-3} 
& GPT-5-mini & 55.3 ($\uparrow$62.6\%) \\
\hline
\end{tabular}
\caption{Accuracy on the \textit{Celebrity 50} benchmark under three settings. Parentheses show relative change vs. baseline.}
\label{tab:main}
\end{table}
\subsection{Evaluation Objectives}
The central goal of our evaluation is to assess whether Bazi-based symbolic features improve a model's ability to fit real-world life events. 
Specifically, we aim to answer three questions:  
(1) Do Bazi-derived features provide incremental information beyond raw birth dates?  
(2) Can hybrid symbolic–LLM models outperform baseline LLMs on reasoning about biographical events?  
(3) Are improvements robust to input perturbations such as shuffling birth dates?

\subsection{Experimental Design}
We design a multiple-choice QA benchmark where each persona is represented by a set of biographical questions grounded in real-life events. 
The input to the model includes the individual's birth date, time, gender, and place of birth, along with a question and candidate answers. 
The task is to select the correct answer, which requires reasoning across symbolic Bazi rules and external world knowledge.

To systematically evaluate performance, we establish three experimental settings:
\begin{itemize}
    \item \textbf{Vanilla LLM + Bazi (Baseline):} Standard LLMs are provided with Bazi-derived features, without additional reasoning modules.  
    \item \textbf{Vanilla LLM + Bazi Rule Knowledge:} In addition to Bazi features, models are augmented with explicit symbolic knowledge rules.  
    \item \textbf{Our Model:} A multi-agent architecture that integrates symbolic reasoning and LLM inference for Bazi-inspired character simulation.
\end{itemize}

To validate the importance of birth-date grounding, we introduce a \textbf{Shuffled Birthday Control}. 
In this condition, each subject's true birth date is replaced with another person's date while keeping all other information constant. 
If Bazi reasoning is meaningful, performance should deteriorate when the mapping between biography and true birth time is shuffled. Finally, we evaluate our model’s performance on the questions from The 15th Global Fortune-Teller Championship 2024 organized by the Hong Kong Junior Feng Shui Masters Association\footnote{We also released the formatted datasets from the Global Fortune-Teller Championship covering the years 2010$–$2024 in our project repository, which will be continuously updated.}
\cite{Wang2024FortuneCompetition}.
\subsection{Implementation Details}
\begin{table}[ht]
\centering
\small
\begin{tabular}{|l|l|c|}
\hline 
Setting & Model & Acc. (\%) \\
\hline 
\multirow{3}{*}{Real Birthdays} 
 & DeepSeek-V3 & 51.2 \\
 & Gemini-2.5-flash & 47.1 \\
 & GPT-5-mini & 55.3 \\
\hline 
\multirow{3}{*}{Shuffled Birthdays} 
 & DeepSeek-v3 & 40.6 ($\downarrow$20.7\%) \\
 & Gemini-2.5-flash & 35.5 ($\downarrow$24.6\%) \\
 & GPT-5-mini & 30.0 ($\downarrow$45.7\%) \\
\hline
\end{tabular}
\caption{Accuracy (\%) on the \textit{Celebrity 50} benchmark under our proposed \textbf{BaZi reasoning model}. 
The shuffled condition replaces each sample's birthday with another person's, breaking the correspondence between real birth dates and BaZi features. 
The significant accuracy drop across all models verifies that our BaZi reasoning framework effectively leverages symbolic birth information for character fitting.}
\label{tab:shuffle_our}
\end{table}
We evaluate three representative backbones: DeepSeek-v3, Gemini-2.5-flash, and GPT-5-mini—under all experimental settings.  
The evaluation is conducted on \textbf{Celebrity 50}. Accuracy is reported as the primary metric. For each model, we measure relative performance changes across conditions to quantify the contribution of Bazi features and symbolic reasoning.

\subsection{Results and Analysis}
\begin{table}[ht]
\centering
\small
\begin{tabular}{|l|l|c|}
\hline
Setting & Model & Acc. (\%) \\
\hline
\multirow{3}{*}{Vanilla LLM + Bazi} 
 & DeepSeek-V3 & 39.3 \\
\cline{2-3} & Gemini-2.5-flash & 42.2 \\
\cline{2-3} & GPT-5-mini & 34.0 \\
\hline
\multirow{3}{*}{/ + Shuffled Birthday} 
 & DeepSeek-V3 & 42.5 \,($\uparrow$8.1\%) \\
\cline{2-3} & Gemini-2.5-flash & 42.1 \,($\downarrow$0.2\%) \\
\cline{2-3} & GPT-5-mini & 34.8 \,($\uparrow$2.4\%) \\
\hline
\end{tabular}
\caption{Accuracies (\%) on the \textit{Celebrity 50} benchmark using \textbf{vanilla LLMs}. 
Each model was provided with the BaZi features derived from the subject's real birthday. 
In the shuffled setting, the input BaZi features were replaced with those derived from another person's birthday, while keeping the rest of the setup unchanged. 
Values in parentheses indicate the relative change compared to the real-birthday setting.}
\label{tab:shuffle_bazi}
\end{table}
Table~\ref{tab:main} compares baseline LLMs, rule-augmented variants, and our model. Our hybrid system consistently outperforms baselines, with relative accuracy gains ranging from +6.6\% to +62.6\% across different backbones. Table~\ref{tab:shuffle_our} reports the effect of shuffled birthdays, where performance drops by up to 45.7\%, confirming that genuine temporal alignment between birth data and biographical outcomes is critical for effective reasoning. Table~\ref{tab:shuffle_bazi} further compares shuffled vs.~real birthdays under the vanilla LLM setting, highlighting that without symbolic integration, LLMs fail to exploit BaZi features consistently.

A closer look reveals that vanilla LLMs contain only limited implicit knowledge of BaZi. As a result, their performance remains relatively stable even when birthdays are shuffled, since their reasoning is not strongly grounded in symbolic BaZi features. In contrast, our BaZi reasoning model explicitly encodes and interprets BaZi structure; therefore, it achieves much higher accuracy under the real-birthday condition but suffers a sharper drop when temporal alignment is broken. This contrast indicates that our model is truly leveraging BaZi theory rather than relying on surface-level correlations. Overall, these results demonstrate that integrating BaZi into character modeling not only improves accuracy over mainstream LLMs but also highlights the potential of culturally grounded symbolic frameworks for building more realistic and temporally dynamic virtual personas.

Notably, our model generates BaZi knowledge using DeepSeek-R1 \cite{DeepSeekR1} and performs reasoning with Doubao-1.5-Thinking-Pro\cite{Doubao15Pro}. When tested on the question set from The 15th Global Fortune-Teller Championship 2024, the model achieved an accuracy of 60\%, matching the third-place performance in that year's competition. This result suggests that the model's capability can be further improved when coupled with more powerful reasoning engines.
\section{Case Study}
\label{sec:case_study}

We conducted a comparative analysis of \textbf{DeepSeek-V3}, \textbf{GPT-5-mini}, and \textbf{Gemini-2.5-flash} within our custom BaZi analysis framework using a real-world consultation case (sergey\_brin\_P042). The results highlight key differences in interpretation, reasoning, and output style.

\subsection{Differences in BaZi Theory Interpretation}
At the stage of fundamental theoretical analysis, DeepSeek-V3 and Gemini-2.5-flash both classified the chart as a \textit{Shangguan Structural Pattern (伤官格)}, whereas GPT-5-mini identified it as a \textit{Cong Er Structural Pattern (从儿格)}. This divergence led to opposite conclusions regarding \textit{favorable/unfavorable elements (喜/忌)} and the direction of future luck cycles. While flexibility exists in pattern classification, such decisions typically rely on the experience of a professional consultant. These results suggest that GPT-5-mini adopts a more flexible and bold interpretative logic, while Deepseek-V3 and Gemini-2.5-flash exhibit a more conservative, rule-bound approach.

\subsection{Differences in Scene Mapping Process}
DeepSeek-V3 primarily follows a "feature-to-prediction" pattern, which can appear rigid and more susceptible to local information bias. In contrast, Gemini-2.5-flash integrates multiple dimensions of chart features to form a holistic analysis. GPT-5-mini demonstrates behavior most similar to a human consultant, adapting its reasoning to the user's current life context and exploring alternative scenarios dynamically.

\subsection{Differences in Output Expression}
DeepSeek-V3 often maps BaZi characteristics directly to real-world manifestations using absolute statements. Gemini-2.5-flash and GPT-5-mini, however, employ more probabilistic language (e.g., "possibly", "likely") and present multiple potential outcomes. While DeepSeek-V3 may appear more accurate when its predictions align with reality, it risks losing user trust when predictions fail due to a lack of nuance.

\subsection{Commonalities}
Across all three models, when provided with identical upstream results, the subsequent reasoning paths converge, and no severe factual or logical errors were observed. All models demonstrate a comparable level of baseline BaZi knowledge, sufficient for general consultation purposes. However, none of the models currently exhibit a strong reflection or self-correction mechanism within this analytical framework.

\subsection{Overall Assessment}
For the theoretical reasoning stage, Gemini-2.5-flash provides the most stable and conservative judgments, showing resilience against local noise. GPT-5-mini tends to produce more aggressive and exploratory interpretations, whereas DeepSeek-V3 remains rigid and deterministic. For the final output stage, GPT-5-mini performs best, generating explanations most similar to those of a human consultant.

\section{Related Work}

\subsection{AI-Driven NPC Development in Games}

The application of artificial intelligence to non-player character (NPC) behavior represents a well-established research area with substantial academic coverage. Karaca et al. \cite{karaca_2023_ai_npc} provide a comprehensive analysis of AI-powered procedural content generation for enhancing NPC behavior, examining how deep learning techniques create adaptive and personalized gaming experiences through reinforcement learning and neural network approaches.

\hspace*{1em}Recent systematic reviews by Zeng \cite{zeng_2023_npc} identify key challenges in creating human-like NPC behavior, categorizing AI techniques into planning, user interaction, position modification, parameter modification, character state modification, and target assignment strategies. This work demonstrates significant progress in making NPCs more intelligent and responsive while reducing development complexity.

\hspace*{1em}Kopel \cite{kopel_2018_npc} presents experimental results implementing AI techniques, including decision trees, genetic algorithms, and Q-learning for 3D game NPCs. The research compares different approaches for creating believable character behavior and demonstrates practical applications of machine learning in game development.

\hspace*{1em}Comprehensive surveys \cite{mehta_2025_ai_games} examine AI's role in game development and player experience, highlighting dynamic difficulty adjustment, procedural content generation, and adaptive NPC systems. This work demonstrates how AI enables personalized gameplay through technologies like the Nemesis System in Middle-earth: Shadow of Mordor.

\hspace*{1em}Research on evolutionary algorithms for NPC behavior by Armanto et al. \cite{armanto_2024_npc} provides a systematic analysis of how genetic algorithms can optimize NPC interactions and decision-making processes. The work establishes six categories for evolutionary algorithm applications in NPC development.

\hspace*{1em}Research by Filipović \cite{filipovic_2023_ai_games} examines AI applications in game development, including computational linguistics aspects of how natural language processing enables more sophisticated dialogue systems and character interactions, bridging the gap between artificial and natural language in game contexts.

\hspace*{1em}The field continues to evolve with the integration of large language models for dynamic NPC dialogue generation, though challenges remain in maintaining character consistency and managing computational complexity \cite{wikipedia_ai_games_2025}.

\subsection{Interactive Storytelling and Computational Narratives}

Interactive storytelling represents a mature research field examining how computational systems can generate, manage, and adapt narratives in response to user interaction. Szilas \cite{szilas_2007_narrator} established foundational work on intelligent narrators for interactive drama, proposing rule-based systems that dynamically maintain storylines while adapting to user intervention. This approach models narrative through computational simulation of narrative laws.

\hspace*{1em}Contemporary research by Beguš \cite{begus_2024_experimental} provides a comparative analysis between human-authored and AI-generated stories, examining 250 human-created and 80 AI-generated narratives. The research reveals that while large language models produce structurally coherent stories, they struggle with emotional authenticity and psychological complexity.

\hspace*{1em}Kybartas and Bidarra \cite{mdpi_2023_survey} present comprehensive surveys of computational and emergent digital storytelling, analyzing bottom-up emergent narratives versus top-down drama manager approaches. Their work examines how AI integration elevates NPC interactions and creates more immersive narrative experiences.

\hspace*{1em}Recent developments in narrative frameworks by Gerba \cite{gerba_2025_narrative} propose Universal Narrative Models for computational storytelling, addressing the "player dilemma" between narrative coherence and user agency. These frameworks separate storytelling from narrative structure to enable greater creative flexibility while maintaining coherent progression.

\hspace*{1em}Research on narrative intelligence and cultural transmission by Cavazza et al. \cite{cavazza_2003_interactive} examines how AI systems can understand and respond to stories, exploring the connection between AI formalisms and narrative analysis. This work addresses challenges in authoring interactive narratives and managing user freedom within structured story frameworks.

\hspace*{1em}Studies of AI-powered narrative generation by Kabashkin et al. \cite{kabashkin_2025_archetypal} investigate how large language models reproduce archetypal storytelling patterns, finding that AI excels at structured, goal-oriented narratives but struggles with psychologically complex and ambiguous stories.

\hspace*{1em}The field increasingly explores hybrid human-AI collaboration in storytelling, examining how computational systems can serve as co-creators while maintaining narrative consistency and emotional depth across extended interactions.
\subsection{Traditional Chinese Metaphysics and Bazi Theory}

The academic study of traditional Chinese divinatory systems, particularly Bazi (八字) or Four Pillars of Destiny, represents a limited but growing field within sinology and anthropological research. Historical research by Pankenier \cite{pankenier_2013_astrology} examines court astrology in late sixth and early seventh century China, including "field allotment" divination and mantic responses to astrological events. This work demonstrates how astrological practices were integrated into imperial governance and policy-making decisions.

\hspace*{1em}Mak \cite{mak_2014_yusi} provides valuable insight into the transmission of Western astral science into Chinese contexts through analysis of the Yusi jing (聿斯經), examining how Hellenistic astrological concepts were adapted and integrated into Chinese divinatory traditions. The research reveals the complex intercultural exchange that shaped medieval Chinese astrology.

\hspace*{1em}Contemporary ethnographic work by academic contributors \cite{academia_chinese_medicine_2013} explores the relationship between Chinese astrology and traditional Chinese medicine, illustrating how birth date analysis through the "eight characters" system connects to health assessment and personality characterization within TCM frameworks. However, the field lacks a comprehensive peer-reviewed analysis of Bazi's epistemological foundations and contemporary applications.

\section{Future Directions}
In terms of the model, future improvements should focus on two areas: Incorporating domain-specific knowledge bases or training for particular schools of BaZi thought (e.g., specialized pattern classifications); Implementing agent-based mechanisms that can dynamically select among intermediate outputs, reflect on user feedback, and adapt reasoning pathways accordingly. For the dataset, we will collect data samples from different countries with more precise birth times. 

\section{Conclusion}
We introduced the first QA dataset and the first LLM system for \textit{BaZi-based persona reasoning}. By integrating symbolic BaZi features with LLMs, our approach enables fine-grained and temporally dynamic character simulation. Experiments on the \textit{Celebrity 50} benchmark show significant accuracy gains over mainstream LLMs, highlighting the value of culturally grounded symbolic–LLM integration for realistic persona modeling.

\section*{Limitations}
While the Celebrity~50 dataset provides a useful benchmark, several limitations remain: (1) many narratives and questions are LLM-generated (by Qwen), introducing potential hallucination, bias, and factual errors; (2) the dataset is small (50 individuals, 488 questions) and gender-imbalanced (37 male, 13 female), limiting generalizability; (3) birth details, though sourced from astro.com, may still contain inaccuracies; and (4) the focus on mostly Western figures born around 1940 introduces temporal and cultural biases that may not generalize across eras or contexts.

\section*{Acknowledgements}
This work was supported by the MirrorAI Fund. We would like to thank Cunxiang Wang, Heguang Lin, Yulong Chen, Jenny, Yinpeng Ma and all reviewers for their valuable feedback and insightful suggestions on this research. We also gratefully acknowledge Prof. Shengzhong Xiao (Wuhan University) for providing verified birth data of notable figures used in our dataset validation.

\bibliography{custom}

\appendix

\end{CJK}
\end{document}